\documentclass[sigconf]{acmart}

\usepackage{multirow} 
\usepackage{balance}

\usepackage[marginal]{footmisc}
\AtBeginDocument{%
  }

\copyrightyear{2026}
\acmYear{2026}
\setcopyright{cc}
\setcctype{by}
\acmConference[WWW '26]{Proceedings of the ACM Web Conference 2026}{April 13--17, 2026}{Dubai, United Arab Emirates}
\acmBooktitle{Proceedings of the ACM Web Conference 2026 (WWW '26), April 13--17, 2026, Dubai, United Arab Emirates}
\acmPrice{}
\acmDOI{10.1145/3774904.3793032}
\acmISBN{979-8-4007-2307-0/2026/04}

\begin{document}

\begin{abstract}
The rapid growth of video content on platforms such as TikTok and YouTube has intensified the spread of multimodal hate speech, where harmful cues emerge subtly and asynchronously across visual, acoustic, and textual streams. Existing research primarily focuses on video-level classification, leaving the practically crucial task of temporal localisation, identifying when hateful segments occur, largely unaddressed. This challenge is even more noticeable under weak supervision, where only video-level labels are available, and static fusion or classification-based architectures struggle to capture cross-modal and temporal dynamics.
To address these challenges, we propose MultiHateLoc, the first framework designed for weakly-supervised multimodal hate localisation. MultiHateLoc incorporates (1) modality-aware temporal encoders to model heterogeneous sequential patterns, including a tailored text-based preprocessing module for feature enhancement; (2) dynamic cross-modal fusion to adaptively emphasise the most informative modality at each moment and a cross-
modal contrastive alignment strategy to enhance multimodal
feature consistency; (3) a modality-aware MIL objective to identify discriminative segments under video-level supervision.
Despite relying solely on coarse labels, MultiHateLoc produces fine-grained, interpretable frame-level predictions.
Experiments on HateMM and MultiHateClip show that our method achieves state-of-the-art performance in the localisation task. \href{https://github.com/mmilabuk/multihateloc}{Code is available at https://github.com/Multimodal-Intelligence-Lab-MIL/MultiHateLoc}.

{\color{red}Disclaimer: This paper contains sensitive content that may be
disturbing to some readers}
\end{abstract} 


\title{MultiHateLoc: Towards Temporal Localisation of Multimodal Hate Content in Online Videos}





\author{Qiyue Sun}
\authornote{Equal contribution.} 
\authornote{Work was conducted while Qiyue Sun was a Research Assistant at the University of Exeter and a visiting PhD student at the University of Birmingham, with primary affiliation to Shandong University.}

\affiliation{%
  \institution{Department of Computer Science, University of Exeter}
  \country{Exeter, UK}}
\email{julysun@mail.sdu.edu.cn}

\author{Tailin Chen}
\authornotemark[1] 
\affiliation{%
  \institution{Department of Computer Science, University of Exeter}
  \country{Exeter, UK}}
\email{T.Chen2@exeter.ac.uk}

\author{Yinghui Zhang}
\affiliation{%
  \institution{Department of Computer Science, University of Exeter}
  \country{Exeter, UK}}
\email{yz949@exeter.ac.uk}

\author{Yuchen Zhang}
\affiliation{%
  \institution{Institute for Analytics and Data Science, University of Essex}
  \country{Essex, UK}}
\email{yuchen.zhang@essex.ac.uk}

\author{Jiangbei Yue}
\affiliation{%
  \institution{Department of Computer Science, University of Exeter}
  \country{Exeter, UK}}
\email{J.Yue@exeter.ac.uk}

\author{Jianbo Jiao}
\affiliation{%
  \institution{School of Computer Science, University of Birmingham}
  \country{Birmingham, UK}}
\email{j.jiao@bham.ac.uk}

\author{Zeyu Fu}
\authornote{Corresponding author.} 
\affiliation{%
  \institution{Department of Computer Science, University of Exeter}
  \country{Exeter, UK}}
\email{Z.Fu@exeter.ac.uk}

\renewcommand{\shortauthors}{Qiyue Sun et al.}




\maketitle



\vspace*{0.6\baselineskip} 
\vspace{-1em}
\section{Introduction}
\begin{figure}[t]
  \centering
  \includegraphics[width=0.5\textwidth]
  {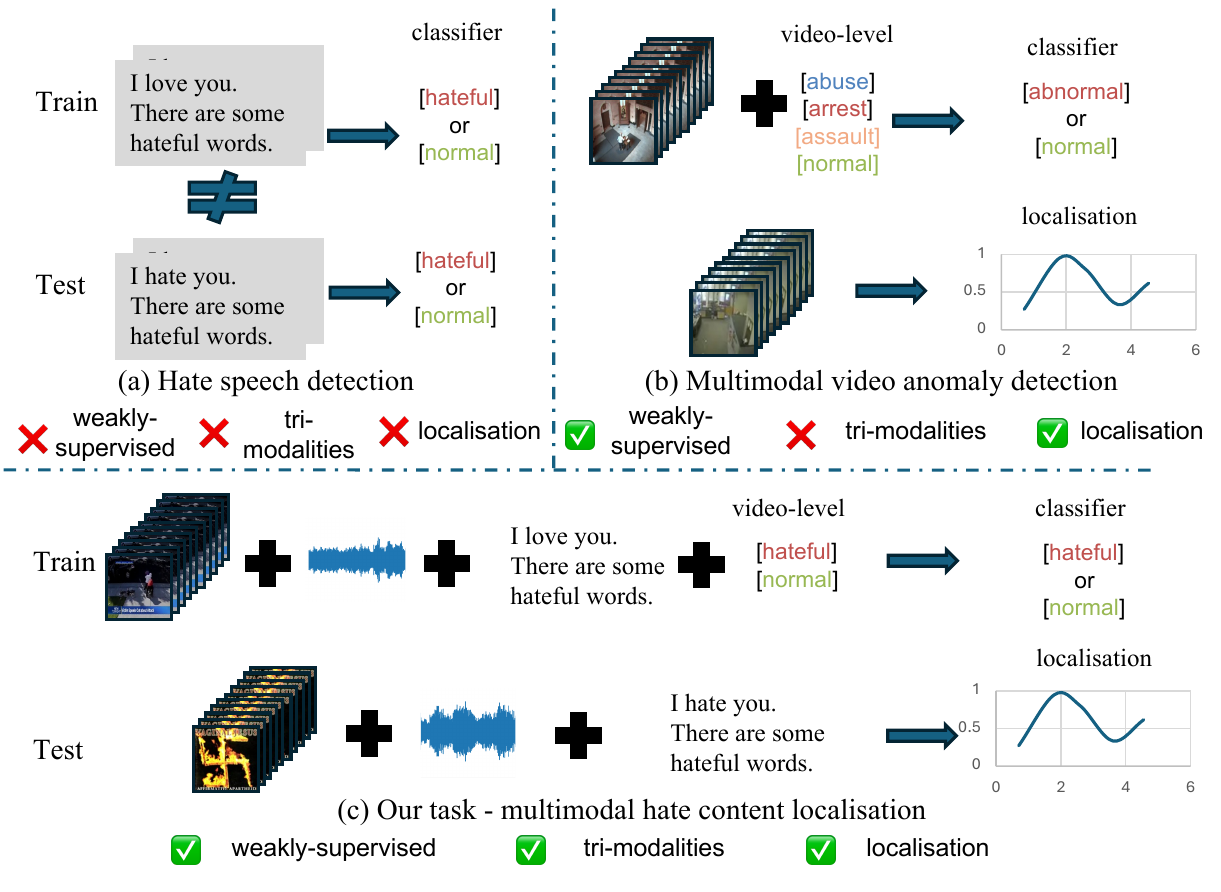} 
  \vspace{-2.5em}
   \caption{Comparison of (a) Hate video detection, (b) Weakly-Supervised multimodal video anomaly detection, and (c) Multimodal hate video localisation.}

  \label{fig:1_ad}
  \vspace{-2em}
\end{figure}
Video content has become the dominant medium for information dissemination on social platforms, with TikTok uploading ~3.2 billion videos in 2021 and YouTube processing 500 hours of uploads per minute by early 2022.
This unprecedented scale has accelerated the spread of hate content—defined as communication attacking groups based on race, gender, religion, or other identities. As online video ecosystems continue to grow, platforms face significant challenges in identifying such content due to its implicit, multimodal, and temporally sparse nature.

Early hate speech research primarily focused on text-only detection, leveraging linguistic features or deep neural networks to classify short social media posts such as tweets or comments ~\cite{badjatiya2017deep, del2017hate}.
However, harmful content rapidly evolved into visual and multimodal forms. Memes combine imagery and overlaid text to convey coded insults, satire, or hateful insinuations. This shift motivated multimodal hate meme detection (e.g., Facebook Hateful Memes Challenge~\cite{kiela2020hateful}, Hate Speech Detection~\cite{gomez2020exploring}), demonstrating that cross-modal modelling is essential because hateful intent often emerges only when multiple modalities are considered together.

The focus of multimodal learning has expanded toward video-based hate content, where harmful cues unfold across visual frames, spoken language, background audio, and on-screen text~\cite{zhang2025dehate,yue2025multimodal}. 
Early approaches typically rely on partial modalities, including vision-only abusive behaviour recognition~\cite{jiang2023abusive} and text–audio fusion for spoken toxicity analysis~\cite{alves2023audio}.
Building on HateMM~\cite{das2023hatemm}, several recent works further explore multimodal hate video classification~\cite{yang2025revealing}.
A robust fusion study systematically compares modality combinations and feature encoders on HateMM and Hateful Memes~\cite{robustMMHate2025}, CMFusion~\cite{CMFUsion2025} employs channel-wise and modality-wise fusion to strengthen tri-modal representation learning, and MoRe~\cite{lang2025more} leverages retrieval-augmented evidence and a mixture-of-experts design to enhance robustness in short videos, while MM-HSD~\cite{mmhsd2025}
and MultiHateGNN~\cite{yue2025multimodal} introduce stronger multimodal backbones with cross-modal attention or graph neural networks for video-level prediction. ImpliHateVid~\cite{rehman2025impliHateVid} extends to implicit hate in videos and also reports results on HateMM, and few-shot multimodal methods~\cite{ma2025fewshot} address label scarcity on the same dataset.

However,  all these approaches produce a single binary label at the video level, classifying an entire video as either hateful or non-hateful, without identifying where within the video the harmful content actually occurs, as illustrated in Fig.~\ref{fig:1_ad}(a). This highlights a major limitation in current research: the lack of fine-grained temporal localisation. In real-world scenarios, hateful content often appears only in short segments of a video, while the rest may be benign. Simply labelling an entire video as hateful overlooks this nuance and limits the practicality of content moderation.

To address this gap, we study a new research task: multimodal hate content localisation, which aims to identify the precise temporal segments in which hate occurs across modalities, as shown in Fig.~\ref{fig:1_ad} (c). Rather than requiring detailed frame-level annotations, we explore this problem under a weakly supervised setting, inspired by recent advances in weakly supervised video anomaly detection (WSVAD) (see Fig.~\ref{fig:1_ad} (b)). WSVAD methods~\cite{sultani2018real,tian2021weakly,wu2024weakly} are effective in learning temporal patterns from video-level labels by aligning visual and audio cues over time. However, they typically overlook the text modality, which is critical in the context of multimodal hate detection. Therefore, a major challenge in our work is to extend weakly supervised approaches to explicitly incorporate temporal modelling of text features, enabling a more complete and fine-grained understanding of multimodal hate content.
A second major challenge in localising multimodal hate content lies in the high variability of modality relevance across time. Hate cues may not be evenly distributed; text might be dominant in one segment, while audio or visual signals become more salient elsewhere. This context-dependent salience makes it ineffective to apply static or early fusion strategies.

\begin{figure*}[t]
  \centering
  \includegraphics[width=0.9\textwidth]
  {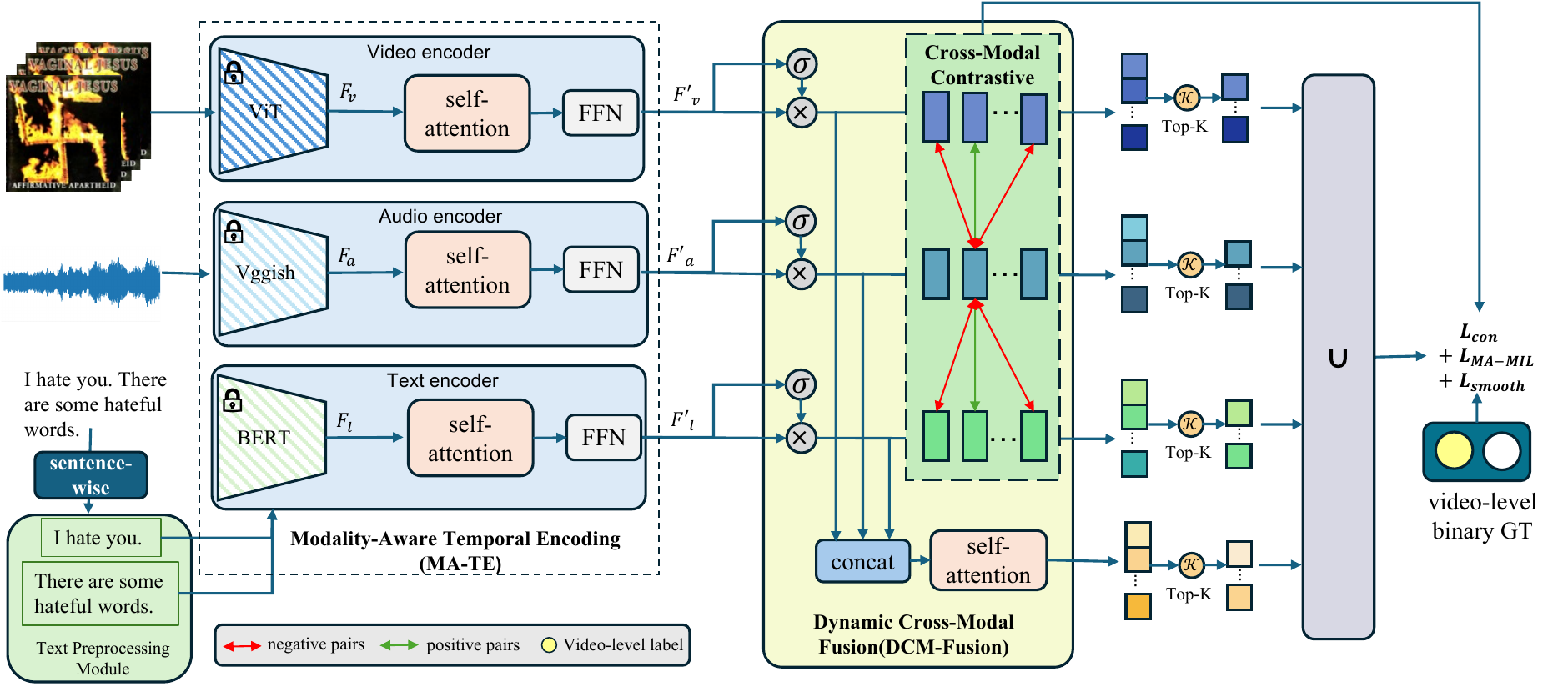} 
   \caption{Overview of the proposed MultiHateLoc framework. The framework integrates multi-modal features from video, audio, and text modalities, leveraging a uniform temporal modeling approach, a novel dynamic cross-modal fusion strategy, and a modality-wise top-K MIL loss. $F_v$, $F_a$, and $F_t$ represent the extracted features from video, audio and text, respectively. $F_m$ indicates features derived from any one of these modalities.
   }
  \label{fig:framework}
  \vspace{-1em}
\end{figure*}

These challenges motivate our design of \textbf{MultiHateLoc}, as shown in Fig~\ref{fig:framework}, the first multimodal hate localisation framework which incorporates 1) Modality-aware temporal modeling to capture intra-modal sequential information across dense video frames, audio features, and text sequences. We further proposed a novel text embedding module for sentence-wise text feature extraction.
2) Dynamic cross-modal fusion to adaptively emphasise the modality contributions based on context at each time-step and a cross-modal contrastive alignment strategy to enhance multimodal
feature consistency. 3) Modality-aware top-K Multiple Instance Learning (MIL) objective that guides weak supervision toward the most relevant segments. Together, these components overcome the inherent limitations of prior work and enable fine-grained localisation of multimodal hate content.
We evaluate MultiHateLoc on the HateMM~\cite{das2023hatemm} and MultiHateClip~\cite{wang2024multihateclip} datasets, achieving state-of-the-art localisation performance. To the best of our knowledge, this is the first work to provide a fully tri-modal, weakly-supervised, and temporally precise solution for hate content localisation in online videos.



Our contributions are summarised as follows:
\begin{enumerate}
    \item We formalise the weakly-supervised multimodal hate content localisation task, distinguishing it from classification and anomaly detection.
    \item We propose MultiHateLoc, which integrates newly proposed temporal modelling, dynamic cross-modal fusion, and modality-aware MIL to enable precise segment predictions.
    \item Extensive experiments on two benchmark datasets validate our approach, establishing new benchmarks for the task.
\end{enumerate}
\section{Related work}

\subsection{Hate Speech Detection}

Early research on hate speech detection has largely centered around textual content, where a wide array of machine learning and deep learning techniques have been explored. For example, MacAvaney et al.~\cite{macavaney2019hate} employed a multi-view SVM approach that combines diverse feature representations to improve both accuracy and interpretability. Badjatiya et al.~\cite{badjatiya2017deep} investigated a range of neural models, including FastText, CNNs, and LSTMs, demonstrating their effectiveness on social media datasets such as tweets. Other conventional techniques, such as support vector machines, have also been utilized for hate speech classification~\cite{del2017hate,sanoussi2022detection}.

Despite promising results, these approaches are fundamentally limited by their reliance on a single modality — text. In real-world online discourse, hate speech can be highly nuanced, leveraging sarcasm, coded language, or context-dependent cues that are difficult to capture with text alone. This makes purely textual models susceptible to false negatives, particularly when hateful intent is implied rather than explicit~\cite{yin2021towards,rottger2020hatecheck,grondahl2018all}. These limitations underscore the need for models capable of integrating richer contextual and multimodal signals to improve robustness and generalizability.

\subsection{Multimodal Hateful Content Detection}

Multimodal approaches substantially outperform unimodal systems for hateful content detection. Early multimodal works combined image–text or video–text signals using concatenation, gating, or bilinear pooling~\cite{yang2019exploring,gomez2020exploring,chhabra2023multimodal}, and transformer-based architectures such as ViLBERT further improved cross-modal alignment~\cite{kiela2020hateful}. More recent efforts extend to tri-modal settings, with HateMM~\cite{das2023hatemm} demonstrating the benefits of jointly modeling video, audio, and text.  
Two main directions have emerged: (1) adapting pre-trained multimodal models, e.g., CLIP-based systems like VAD-CLIP~\cite{wu2023vadclip}, which incorporate temporal modeling but still rely on global text features; and (2) task-specific fusion architectures such as MM-HSD~\cite{cespedes2025mm}, CMFusion~\cite{CMFUsion2025}, MoRE~\cite{lang2025biting} MultiHateGNN~\cite{yue2025multimodal}. More recent works such as \cite{yang2025revealing} explore the label noise in hate videos and in DeHate\cite{zhang2025dehate}, a novel fine-grained large-scale hate content dataset has been proposed for advanced hate detection. While these methods advance video-level classification, a recent survey~\cite{hee2024recent} highlights persistent temporal misalignment across modalities and the lack of fine-grained grounding, especially for text and audio cues. Consequently, existing multimodal hate detection methods remain classification-centric, leaving the problem of temporal localisation largely unexplored.

\subsection{Weakly Supervised Video Anomaly Detection}
Frame-level annotations are costly to obtain, so weakly supervised learning has become a common solution for temporal localisation tasks. Multi-Instance Learning (MIL) is widely used in this setting. It trains models with only video-level labels and allows them to infer the segments that contain target events. MIL-based methods have shown strong performance in action recognition and video anomaly detection. Examples include the MIL ranking framework by Sultani et al.~\cite{sultani2018real}, self-supervised anomaly modeling by Rudolph et al.~\cite{Rudolph_2021_WACV}, and GNN-based approaches for group-level anomalies~\cite{ullah2023ad,li2022weakly}. These works demonstrate that weak supervision can capture temporal patterns without detailed annotations. However, most research focuses on explicit visual anomalies in surveillance videos, which differs from the subtle and often cross-modal nature of hateful content.

Hate signals are frequently concealed. A video may look visually normal but contain harmful cues in the audio or text. Existing WS-VAD methods~\cite{sultani2018real,li2022weakly,wu2023vadclip} rely mainly on visual features, so they struggle to capture such multimodal patterns. To address this limitation, we propose a weakly supervised MIL framework that integrates video, audio, and textual features. It uses only video-level labels but can still predict the temporal boundaries of hateful segments. This approach enables fine-grained localisation under weak supervision and fills an important gap in applying WS-VAD techniques to multimodal hate content.

\section{Methodology}
To address the need for weakly supervised multimodal hate localisation (WS-MHL), we introduce MultiHateLoc, a framework designed for detecting hateful segments in videos using only video-level labels. We first define the WS-MHL task and describe its input–output setting. We then explain how each component of MultiHateLoc is designed to meet the requirements of this task. The overall architecture is shown in Fig.~\ref{fig:framework}.

\subsection{Problem Formulation}
\begin{figure}[t]
  \centering
  \includegraphics[width=\columnwidth]{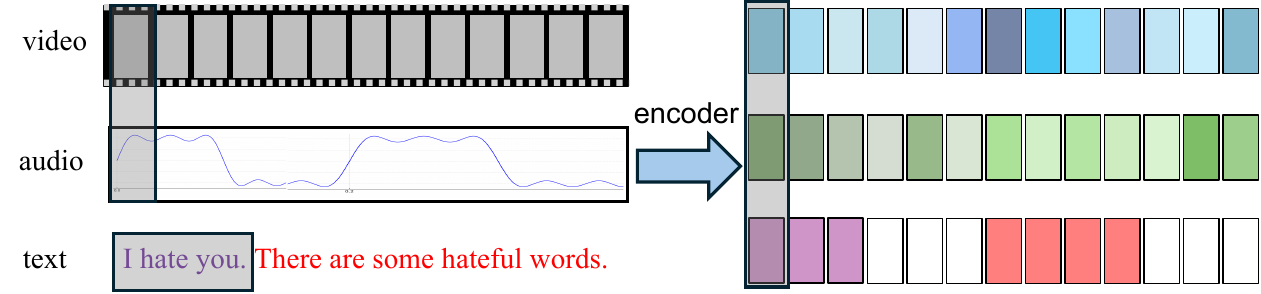}
  \vspace{-1.5em}
   \caption{\textcolor{black}{Sentence-wise} text embedding for text-based localisation. The input video is segmented into aligned video frames, audio signals, and sentence-level text units. Each sentence is encoded into an embedding and propagated to the corresponding temporal regions across video and audio. 
   }
  \label{fig:text_preprocessing}
  \vspace{-2em}
\end{figure}

The \textbf{MultiHateLoc} focuses on temporally localising hate content in videos using only video-level annotations. For a given video \( V \), we extract features from three modalities: video (\( F_v \)), audio (\( F_a \)), and text (\( F_t \)). The goal is to predict frame-level probabilities \( P = \{\hat{y}_1, \hat{y}_2, \dots, \hat{y}_T\} \), where \( T \) is the number of frames, while training on video-level binary labels \( y \in \{0, 1\} \), where \( y = 0 \) indicates non-hate content and \( y = 1 \) indicates hate content. Specifically,

\textbf{Input:} 
\begin{itemize}
            \item Video (\( F_v \in \mathbb{R}^{T \times 768} \)): We extract frame-level visual features using a pre-trained ViT-B/16~\cite{dosovitskiy2020image}. Each frame is processed independently to retain spatial details.
            \item Audio (\( F_a \in \mathbb{R}^{T \times 128} \)): Audio features are extracted from 1-second clips using VGGish~\cite{hochreiter1997long}. The resulting sequence is linearly interpolated to match the video length \( T \). This produces a temporally aligned audio representation with the same frame-level resolution as the visual stream.
            \item Text (\( F_l \in \mathbb{R}^{T \times 768} \)): We propose a novel \textbf{sentence-wise text embeddings }for text-based localisation, {\color{black} as shown in Fig.~\ref{fig:text_preprocessing}}, which generated via four steps: (1) Transcribe video audio to text using Whisper~\cite{radford2023robust} ; (2) Split transcribed text into \( S \) sentence-wise fragments based on each sentence start and end timestamps; (3) Encode each sentence with BERT~\cite{devlin2018bert} to obtain a 768-dimensional feature; (4) Expand sentence-level features to the frame length \( T \) by repeated padding within each sentence’s timestamp interval. This converts \( S \) sentence embeddings into frame-aligned text features and resolves temporal misalignment in naive text encodings.
            \item Supervision Signal: We use only video-level binary labels $y \in \{0,1\}$. A label $y=1$ indicates that the video contains at least one hateful segment with a duration of at least $1$ second, while $y=0$ indicates the absence of hateful content.
\end{itemize}
\textbf{Output:} 
The model learns a function \( C \) that maps multi-modal features to {frame-level probabilities}:  A frame-level hate probability sequence \( P = \{\hat{y}_1, \hat{y}_2, \dots, \hat{y}_T\} \) where \( \hat{y}_t \in [0,1] \) (probability that frame \( t \) contains hate content). Probabilities are generated via sigmoid activation to ensure range consistency.
The predictions are evaluated using {frame-level mAP}, which assesses the precision-recall relationship over all frames in the dataset.

\subsection{Modality-Aware Temporal Encoding (MA-TE)}

To fully exploit the unique characteristics and temporal patterns inherent in each modality, our framework employs a \textbf{modality-aware temporal encoding} module, as introduced in Fig.~\ref{fig:framework} (MA-TE), for each input stream (video, audio, and text). Rather than applying a shared or unified temporal backbone, we design dedicated Transformer-based encoders tailored to each modality’s feature space and temporal dynamics.



To capture temporal dependencies unique to each modality, we employ a dedicated Transformer-based temporal encoder for every stream (\(m \in \{v, a, l\}\)). Given extracted features \(F_m \in \mathbb{R}^{T \times D_m}\), we apply self-attention and feed-forward layers as follows:
\begin{equation}
\resizebox{.9\columnwidth}{!}{$
\begin{aligned}
Q_m &= W_q^{(m)} F_m, \quad K_m = W_k^{(m)} F_m, \quad V_m = W_v^{(m)} F_m, \\
F_m' &= \mathrm{FFN}^{(m)}\left(\mathrm{softmax}\left(\frac{Q_m K_m^\top}{\sqrt{d_k}}\right) V_m\right) + F_m
\end{aligned}
$}
\label{eq:modality_specific_encoding}
\end{equation}

where all weights are modality-specific and learned independently. The FFN consists of two fully connected layers with ReLU activation.

This modality-specific temporal encoder ensures that each modality’s sequential patterns are effectively captured and encoded into \(F_m' \in \mathbb{R}^{T \times D}\), facilitating robust and consistent representation for downstream multimodal fusion.

\subsection{Cross-Modal Contrast (CM-Contrast)}
Hateful content localisation poses unique challenges: hate cues are fleeting, weakly labeled, and often distributed unevenly across modalities. To address this, we introduce a \textbf{cross-modal contrastive loss} that enforces temporal alignment between video, audio, and text streams (Details are shown in Fig.~\ref{fig:framework} (CM-Contrast)). This mechanism is not only theoretically appealing but also practically necessary, as it amplifies subtle or modality-specific hate signals and enables robust localisation under weak supervision.

For each time step $t$, let $\mathbf{f}_t^m$ denote the normalised feature embedding of modality $m \in \{v, a, l \}$ (video, audio, text). Our goal is to explicitly maximize the similarity between all pairs of modalities at the same timestamp (positive pairs), while separating them from features sampled at mismatched timestamps or from different videos in the batch (negative pairs). The contrastive alignment mechanism is illustrated in ~Fig.\ref{fig:framework}, where positive pairs (green) are attracted, and negative pairs (red) are pushed away in the shared embedding space.
The proposed contrastive loss is:
\begin{equation}
\resizebox{.9\columnwidth}{!}{$
\mathcal{L}_{\text{con}} = -\frac{1}{|P|} \sum_{(m,n)\in P} \, \mathbb{E}_{(i,t)} \bigg[
    \log \frac{ \exp\left( \frac{ \mathrm{sim}(\mathbf{f}_t^{m,i}, \mathbf{f}_t^{n,i}) }{ \tau } \right) }
    { \sum_{(j,s)} \exp\left( \frac{ \mathrm{sim}(\mathbf{f}_t^{m,i}, \mathbf{f}_s^{n,j}) }{ \tau } \right) }
\bigg]$}
\end{equation}
where \(P = \{(v,a), (a,l), (v,l)\}\) is the set of modality pairs. Here, \(\mathrm{sim}(\cdot,\cdot)\) denotes cosine similarity, and \(\tau\) is a temperature parameter. $i$ indexes the video in the batch, $t$ denotes the current timestamp of video $i$, $j$ indexes all videos in the batch (including $i$ itself), and $s$ indexes all timestamps in the corresponding video. The numerator corresponds to the positive pair $(m,n)$ from the same video and timestamp, while the denominator sums over all positive and negative pairs across videos and timestamps.



\subsection{Dynamic Cross-Modal Fusion (DCM-Fusion)}

\begin{figure}[t]
  \centering
  \includegraphics[width=\columnwidth]{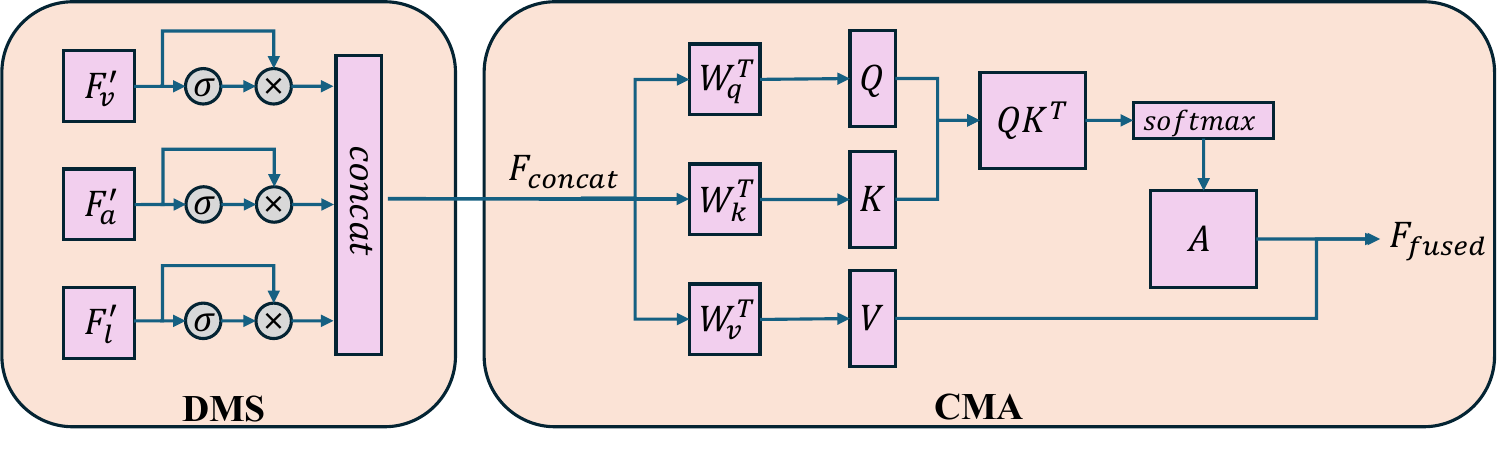}
  \vspace{-2em}
   \caption{Overview of Dynamic Cross-Modal Fusion.
   }
  \label{fig:Overview of Dynamic Cross-Modal Fusion}
  \vspace{-2.5em}
\end{figure}

To integrate features across modalities, as shown in Fig.\ref{fig:Overview of Dynamic Cross-Modal Fusion}, we propose a \textbf{Dynamic Cross-Modal Fusion} strategy, which dynamically adjusts modality contributions at each time step and captures inter-modal interactions.

\noindent \textbf{Dynamic Modality Selection (DMS).} For each time step \( t \), modality-specific importance weights are dynamically computed:
\begin{equation}
\alpha_m[t] = \sigma(W_\alpha^m \cdot F_m'[t] + b_\alpha^m), \quad m \in \{v, a, l\},
\end{equation}
where \( \sigma \) is the Sigmoid activation function. These weights reweight the modality-specific features:
\begin{equation}
F_m^{\text{weighted}}[t] = \alpha_m[t] \cdot F_m'[t].
\end{equation}

Such a strategy ensures that the framework remains robust to noisy or missing modalities by dynamically adjusting their contributions based on the context at each time step.

\noindent \textbf{Cross-Modal Attention (CMA).} The reweighted modality features are concatenated and processed using a cross-modal attention mechanism to capture inter-modal dependencies. The concatenated feature is:
\begin{equation}
\resizebox{.9\columnwidth}{!}{$
F_{\text{concat}} = \text{Concat}(F_v^{\text{weighted}}, F_a^{\text{weighted}}, F_l^{\text{weighted}})
\in \mathbb{R}^{T \times 3D}
$}
\end{equation}
From this concatenated feature, Query, Key, and Value matrices are generated via learnable transformations:
\begin{equation}
\resizebox{.9\columnwidth}{!}{$
Q = W_q \cdot F_{\text{concat}}, \quad
K = W_k \cdot F_{\text{concat}}, \quad
V = W_v \cdot F_{\text{concat}}
$}
\end{equation}
where \( W_q, W_k, W_v \in \mathbb{R}^{3D \times D} \) are learnable weight matrices.
The fused cross-modal feature is computed as:
\begin{equation}
F_{\text{fused}} = \text{softmax}\left(\frac{Q K^\top}{\sqrt{d_k}}\right) V \in \mathbb{R}^{T \times D}.
\end{equation}





\subsection{Modality-aware Multi-instance Learning (MA-MIL) }

To effectively identify the most informative frames while ensuring balanced contributions from different modalities, we propose a Modality-aware MIL learning strategy. As shown in Fig.~\ref{fig:framework}, with the outputs from DCM-Fusion module, this strategy combines predictions from individual modalities and fused features, addressing the challenges of the weakly-supervised setting.

\noindent 
\textbf{Single-Modality Top-K Selection.}
For each modality, frame-level probabilities \( P_m \in \mathbb{R}^T \) are generated from modality-specific branches, and then the Top-K frames are selected based on their probabilities:
\begin{equation}
\begin{aligned}
P_m = \sigma(W_m \cdot F_m' + b_m), \quad m \in \{v, a, l\}
\\
S_m = \text{Top-K}(P_m), \quad S_m \subseteq \{1, \dots, T\}.
\end{aligned}
\end{equation}
\noindent 
\textbf{Multi-Modal Fusion Top-K Selection.}
Frame-level probabilities from fused features \( P_{\text{fused}} \in \mathbb{R}^T \) are generated and the Top-K frames are selected out:
\begin{equation}
\begin{aligned}
P_{\text{fused}} = \sigma(W_{\text{cls}} \cdot F_{\text{fused}} + b_{\text{cls}})
\\
S_{\text{fused}} = \text{Top-K}(P_{\text{fused}}).
\end{aligned}
\end{equation}

\noindent \textbf{Final Top-K Loss.}
The final set of frames \( S_{\text{final}} \) is constructed by merging the fused Top-K results and modality-specific Top-K results:

\begin{equation}
S_{\text{final}} = S_{\text{fused}} \cup \bigcup_{m \in \{v, a, l\}} \left(w_m \cdot S_m \right),
\end{equation}
where \( w_m \) represents modality-specific importance weights derived from Dynamic Modality Selection. 
Note that \( w_m \cdot S_m \) rescales the probabilities for each modality, effectively reducing the contribution of less important modalities to the final selection. The MIL loss is computed as:
\begin{equation}
\mathcal{L}_{\text{MA-MIL}} = - \frac{1}{K} \sum_{i \in S_{\text{final}}} \log(\hat{y}_i).
\end{equation}
To encourage consistency across frame-level predictions, we introduce a smoothness regularization term:
\begin{equation}
\mathcal{L}_{\text{smooth}} = \frac{1}{T-1} \sum_{t=1}^{T-1} \| \hat{y}_{t+1} - \hat{y}_t \|^2.
\end{equation}
The final loss function combines the MIL loss, smoothness regularization, and the modality contrastive loss:
\begin{equation}
\mathcal{L}_{total} = \mathcal{L}_{\text{MA-MIL}} + \lambda_{\text{smooth}} \mathcal{L}_{\text{smooth}} + \lambda_{\text{con}} \mathcal{L}_{con},
\end{equation}
where \( \lambda_{\text{smooth}} \) and \( \lambda_{\text{con}} \) balance these three components. 





\section{Experiments}
\subsection{\textbf{Datasets and Implemetation Details.}}
We utilise two multimodal datasets: HateMM\cite{das2023hatemm} and MultiHateClip\cite{wang2024multihateclip}, both of which contain video, audio, and text. HateMM includes ~43 hours of videos from BitChute, labeled as “hateful” or “non-hateful.” For hateful videos, frame-level annotations mark specific segments. The dataset contains 431 hateful and 652 non-hateful videos, supporting both classification and localization tasks.
MultiHateClip contains 2,000 videos from YouTube and Bilibili, evenly divided into hateful, offensive, and normal categories. Each video has segment-level labels and additional metadata (e.g., target group, modality, audience). It supports multilingual and fine-grained hate detection.

\noindent \textbf{Evaluation metrics.} We adopt the following metrics to evaluate the proposed framework: (1) Frame-level Mean Average Precision (mAP), which measures precision-recall performance at the frame level; and (2) Area Under the Curve (AUC), which captures overall precision-recall trade-offs across varying thresholds. 

\noindent \textbf{Implementation details.} 
\textcolor{black}{Our model is implemented by PyTorch and optimized using Adam with an initial learning rate of 1e-4. We train with a batch size of 32 for 100 epochs.} The loss weights are set to $\lambda_{\text{smooth}} = 0.1$ and $\lambda_{\text{con}} = 0.2$.




\begin{table}[t]
\centering
\caption{Performance comparison (mAP ↑, AUC ↑) of \textbf{MultiHateLoc} and baselines using different modality combinations on HateMM and MultiHateClip. We evaluate weakly-supervised anomaly detection (VAD-CLIP) and traditional fusion methods (Early Fusion, Late Fusion) under single-modality (V, A, T), dual-modality (V+T, V+A, T+A), and tri-modality (V+A+T) settings. All baseline results are reproduced using the same feature extraction and training pipeline. Best results are shown in \textbf{bold}.}
\vspace{-1em}
\setlength{\tabcolsep}{3pt}
\begin{tabular}{c c c c}
\toprule
\textbf{Model} & \textbf{Modality} & \textbf{HateMM} & \textbf{MultiHateClip} \\
 & & mAP / AUC & mAP / AUC \\
\midrule

\multirow{6}{*}{VAD-CLIP ~\cite{wu2023vadclip}} 
& Video & 0.531 / 0.740 & 0.348 / 0.605 \\
& Audio & 0.563 / 0.762 & 0.405 / 0.650 \\
& Text & 0.498 / 0.712 & 0.367 / 0.610 \\
& V+A & 0.546 / 0.752 & 0.380 / 0.623 \\
& V+T & 0.559 / 0.761 & 0.393 / 0.631 \\
& T+A & 0.550 / 0.755 & 0.385 / 0.620 \\
\midrule

Early Fusion & V+A+T & 0.565 / 0.765 & 0.410 / 0.662 \\
Late Fusion & V+A+T & 0.578 / 0.779 & 0.401 / 0.660 \\
\midrule
CMFusion~\cite{CMFUsion2025} & V+A+T & 0.596 / 0.763 & 0.420 / 0.672 \\
\midrule

\multirow{4}{*}{\textbf{Ours}} 
& V+T & 0.612 / 0.775 & 0.404 / 0.663 \\
& V+A & 0.621 / 0.781 & 0.414 / 0.676 \\
& T+A & 0.617 / 0.777 & 0.408 / 0.669 \\
& V+A+T & \textbf{0.645 / 0.799} & \textbf{0.445 / 0.750} \\
\bottomrule
\end{tabular}
\label{tab:baseline_comparison}
\vspace{-2em}
\end{table}

\vspace{-0.4em}
\subsection{Baseline Comparisons}

Our task of weakly supervised multimodal hate content localisation has not been addressed in existing research. Existing multimodal hate detection studies focus exclusively on video-level classification, while related tasks such as video anomaly detection and audio–visual event localisation differ fundamentally in supervision, semantics, and modality structure. This mismatch makes direct comparison with existing benchmarks infeasible. 

Therefore, instead of reusing mismatched benchmarks, we construct baselines from methods that can be reasonably adapted to our weakly-supervised scenario. Our goal is not to exhaust all possible models, but to build a meaningful and fair comparison set that reflects the core components required by our task. Specifically, we evaluate three complementary categories:

\noindent \textbf{Weakly-supervised anomaly localisation (VAD-CLIP~\cite{wu2023vadclip}):}
These models naturally output temporal saliency curves under video-level labels, making them the closest analogue to our weakly-supervised setup. However, they lack the explicit cross-modal modelling and semantic alignment required for hate content.

\noindent \textbf{Traditional multimodal fusion (Early Fusion, Late Fusion):}
These represent what multimodal hate classification systems could achieve when repurposed for localisation. Their limitations—static weighting and the absence of temporal reasoning—highlight the need for explicit temporal and cross-modal modeling.

\noindent \textbf{Modern multimodal classifier (CMFusion~\cite{CMFUsion2025}):}
We include CMFusion to test whether a state-of-the-art tri-modal classifier can be adapted into a temporal localiser via a sliding-window strategy. This baseline allows us to contrast classification-driven fusion with our temporal, modality-aware approach.

Our evaluation thus focuses on a coherent and fair baseline set, consistent with the principles of prior temporal localisation benchmarks: compare only with methods that match the supervision level and can be adapted without unrealistic assumptions.
As illustrated in Table~\ref{tab:baseline_comparison}, our framework consistently surpasses all baselines on both HateMM and MultiHateClip datasets. Specifically, on the HateMM dataset, our final model achieves an mAP of \textbf{0.645} and an AUC of \textbf{0.799}, significantly outperforming the strongest baseline, Late Fusion (mAP: 0.578, AUC: 0.779), by approximately \textbf{6.7\%} in mAP and \textbf{2.0\%} in AUC. On the MultiHateClip dataset, which is notably more challenging due to its inherent complexity, our model attains an mAP of \textbf{0.445} and AUC of \textbf{0.750}.





\begin{table*}
\centering
\vspace{-1em}
\caption{Ablation results on the HateMM dataset using frame-level mAP and AUC.}
\vspace{-0.8em}
\begin{tabular}{@{}lccccccc@{}}
\toprule
\textbf{Model}       & \textbf{MA-TE} & \textbf{CMA} & \textbf{DMS} & \textbf{CM-Contrast} & \textbf{MA-MIL} & \textbf{mAP} & \textbf{AUC} \\ \midrule
Baseline (Early-Fusion)              & -        & -            & -            & -  & -                  & 0.565                & 0.765 \\         
Baseline (Early-Fusion)              & \checkmark        & -            & -      & -      & -                & 0.577                & 0.766 \\         
Baseline (Late-Fusion)              & \checkmark        & -            & -        & -     & -                  & 0.581                & 0.770                 \\ \midrule
MultiHateLoc  & \checkmark        & \checkmark   & -            & -       & -            & 0.601                & 0.783                 \\ 
MultiHateLoc   & \checkmark        & \checkmark   & \checkmark   & -        & -           & 0.615                & 0.785                 \\ 
MultiHateLoc  & \checkmark        & \checkmark   & \checkmark   & \checkmark    & -      & 0.621     & 0.789    \\ 
MultiHateLoc  & \checkmark        & \checkmark   & \checkmark   & \checkmark    & \checkmark       & \textbf{0.645}       & \textbf{0.799}        \\
\bottomrule
\end{tabular}
\vspace{-1em}
\label{tab:ablation_study}
\end{table*}

Additionally, the comparison between modality combinations (\textit{e.g.}, V+A, V+T, T+A) reveals that MultiHateLoc effectively leverages complementary information among modalities, achieving consistently better results than the single or dual modality configurations in VAD-CLIP. This underscores the importance and effectiveness of our dynamic modality selection and cross-modal fusion strategies.
We also compare our model with up-to-data classification model CMFusion~\cite{CMFUsion2025}, regarding to the fusion part, our model also consistently achieves better performance.

Overall, these quantitative improvements validate MultiHateLoc's capability in accurately localising hate content by effectively integrating multimodal information under weak supervision.

\vspace{-0.4em}
\subsection{Ablation Studies}

In this section, we conduct ablation experiments on the \textbf{HateMM} dataset under the weakly-supervised setting to evaluate the impact of each component in our modality-aware learning framework. 


\subsubsection{\textbf{Incremental Ablation Study.} }
Table~\ref{tab:ablation_study} systematically evaluates the contribution of each module proposed in our MultiHateLoc framework by incrementally incorporating these modules into baseline models. 
We begin from a basic early-fusion baseline that directly concatenates modality features without advanced temporal modeling or fusion strategies, achieving an mAP of 0.565 and an AUC of 0.765. The introduction of \textbf{MA-TE} slightly enhances the performance, confirming the benefit of capturing intra-modal temporal dependencies.
Further improvement is observed when shifting from early fusion to late fusion with temporal modeling, resulting in an AP increase to 0.581 and an AUC to 0.770. This improvement highlights the advantage of independently processing each modality before fusion.



Integrating the \textbf{DCM-Fusion}, which combines \textbf{CMA} and \textbf{DMS}, yields a substantial improvement (mAP: 0.615, AUC: 0.785). This demonstrates the effectiveness of adaptively capturing inter-modal dependencies and adjusting modality contributions based on temporal relevance, thereby reducing noise from less informative modalities.  

Adding the \textbf{CM-Contrast} brings further gains (mAP: 0.621, AUC: 0.789) by aligning heterogeneous modality features at the frame level, strengthening the model’s ability to localise subtle cross-modal hate cues. Finally, integrating the \textbf{MA-MIL} module achieves the best performance (mAP: 0.645, AUC: 0.799), highlighting the advantage of adaptive segment selection under weak supervision.  

Overall, the consistent performance gains validate the design of our modality-aware learning framework, where each module complements the others to progressively enhance weakly-supervised multimodal hate localisation.

{\color{black}\subsubsection{\textbf{Text Encoding Strategy}}
This section verifies the effectiveness of our sentence-wise text encoding by comparing it with naive text processing (common in prior works \cite{das2023hatemm,wang2024multihateclip} and text-omitted settings.
\begin{table}[h]
  \centering
  \vspace{-0.7em}
  \caption{Ablation study on Text Encoding Strategy}
    \vspace{-1em}
  \begin{tabular}{lcc}
    \toprule
    \textbf{Text Encoding Strategy} & \textbf{AP} & \textbf{AUC} \\ \midrule
    VAD-CLIP\cite{wu2023vadclip} + W/O Text                        & 0.546      & 0.752      \\
    VAD-CLIP\cite{wu2023vadclip} + Naive Text                        & 0.551      & 0.770      \\
    VAD-CLIP\cite{wu2023vadclip} + Sentence-wise Text                & 0.578       & 0.779        \\ \midrule
     MultiHateLoc + W/O Text                        & 0.621       & 0.781       \\
    MultiHateLoc + Naive Text                      & 0.620       & 0.785      \\
     MultiHateLoc + Sentence-wise Text                   & \textbf{0.645 }      & \textbf{0.799}        \\
 \bottomrule
  \end{tabular}
  \label{tab:text encoding}
  \vspace{-0.7em}
\end{table}
As shown in Table.~\ref{tab:text encoding}, \textbf{sentence-wise text encoding} significantly improves localisation accuracy which validates (1) Our sentence-wise text encoding effectively solves the temporal misalignment problem of naive text processing methods, enabling precise mapping between text signals and video/audio frames; (2) Text modality is irreplaceable for weakly-supervised multimodal hate content localisation (WS-MHL), as it provides key discriminative signals that video and audio alone cannot capture. These findings fully justify the inclusion of sentence-wise text encoding in the MultiHateLoc framework.

\begin{table}
  \centering
    \vspace{-0.5em}
  \caption{Sensitivity analysis for the adaptive $K$ parameter in Top-K MIL Loss, where $K$ determines the portion of selected frames.}
  \vspace{-1em}
  \begin{tabular}{lccc}
    \toprule
    \textbf{K Value} & \textbf{Frames Selected} & \textbf{mAP} & \textbf{AUC} \\ \midrule
    1 (100\%) & All Frames & 0.612 & 0.758 \\ 
    2 (50\%) & Top 50\% & 0.630 & 0.785 \\ 
    3 (33\%) & Top 33\% & \textbf{0.645} & \textbf{0.799} \\ 
    5 (20\%) & Top 20\% & 0.620 & 0.762 \\ \bottomrule
  \end{tabular}
  \label{tab:adaptive_top_k}
  \vspace{-1em}
\end{table}

}
\subsubsection{\textbf{Top-K Selection Sensitivity with Adaptive K}}

    Table~\ref{tab:adaptive_top_k} investigates the sensitivity and effectiveness of the adaptive selection parameter $K$ within our proposed Modality-Aware Multi-Instance Learning (MA-MIL) loss. Unlike conventional approaches that utilize a fixed number of top-scoring frames, our adaptive strategy dynamically selects a proportion of frames, effectively balancing between comprehensive context incorporation and the emphasis on discriminative content.
    The optimal balance is achieved at $K=3$ (selecting the top 33\% of frames), resulting in the highest mAP of 0.645 and AUC of 0.799. Selecting all frames ($K=1$, 100\%) leads to a noticeable performance drop (mAP=0.612) due to the inclusion of irrelevant or noisy frames. Conversely, overly aggressive frame reduction ($K=5$, top 20\%) excessively restricts contextual information, marginally decreasing the mAP to 0.620. 
    
    These findings validate the effectiveness of our adaptive $K$ selection approach, demonstrating its capability in guiding the model to focus on the most informative frames, thus significantly enhancing the accuracy and robustness of hate content localisation.


\begin{figure*}[!t]
  \centering
  \includegraphics[width=0.78\textwidth]{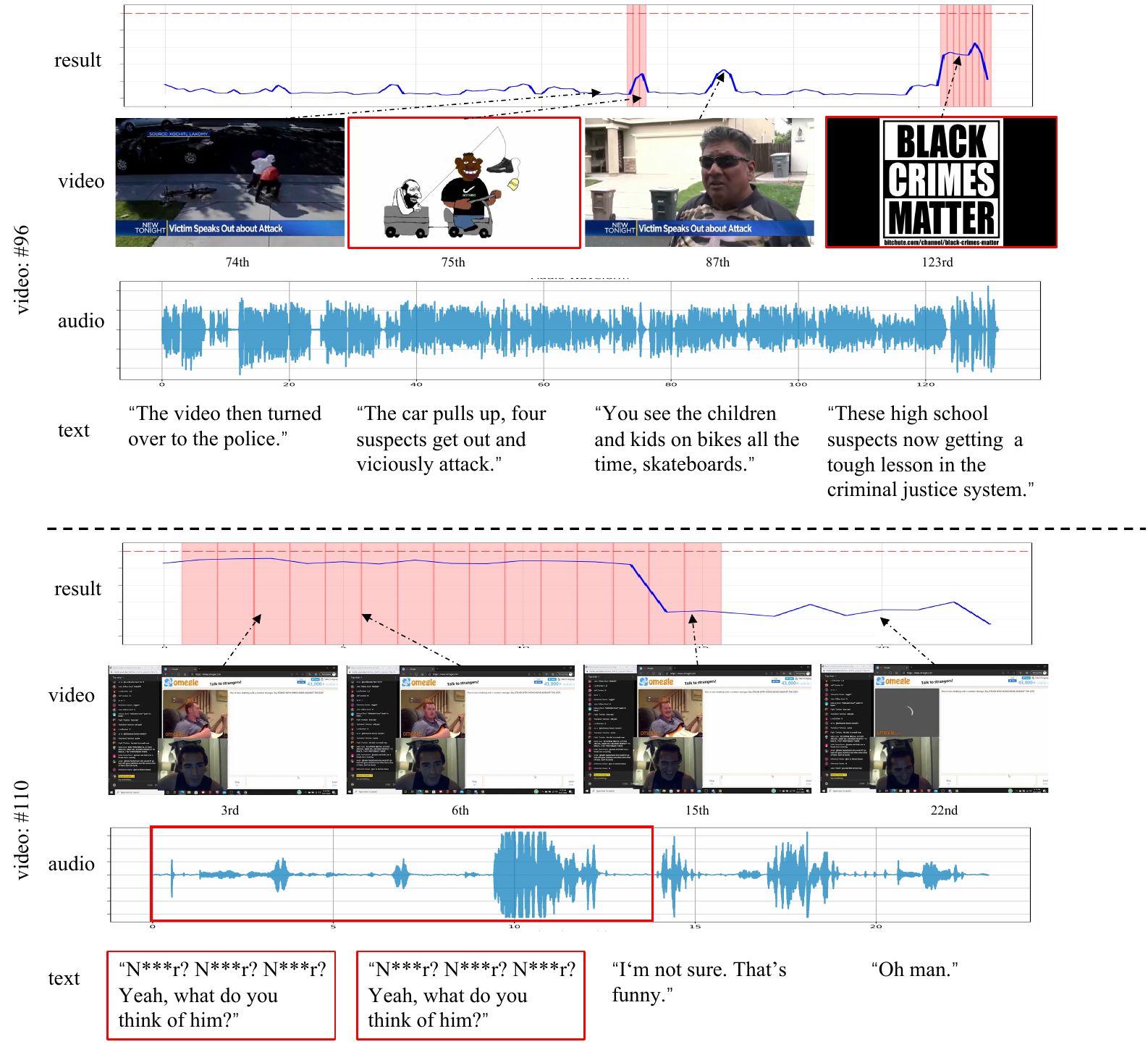}
\caption{Visualization of predictions and modality-specific hate localisation by our proposed \textbf{MultiHateLoc} framework on two samples from the HateMM dataset. For each video, the top curve shows frame-level hate prediction probabilities, where red-shaded areas indicate detected hate segments. Below, temporally aligned frames, audio waveforms, and text transcriptions are shown, with red boxes highlighting modality-specific regions identified as hateful.}
  \label{fig:case_study_visualization}
  \vspace{-1em}
\end{figure*}

\subsubsection{\textbf{Module Simplification}}
Given the complexity of multi-modal fusion, we also explore strategies to simplify individual modules while maintaining performance.

\paragraph{Simplification of Modality Attention}
To balance performance and efficiency, we evaluate the impact of simplifying the modality attention mechanism by varying the number of attention heads. As shown in Table~\ref{tab:modality_attention}, reducing the attention heads from 4 to 2 slightly decreases the AP from 0.635 to 0.630, but significantly reduces the computation cost. Further reduction to a single head causes a noticeable performance drop to 0.620.

\begin{table}[h]
  \centering
  \vspace{-0.7em}
  \caption{Simplification of modality attention by varying the number of attention heads.}
    \vspace{-0.5em}
  \begin{tabular}{ccc}
    \toprule
    \textbf{\#Self-Attention Heads} & \textbf{AP} & \textbf{AUC} \\ \midrule
    1                        & 0.620       & 0.780        \\\midrule
    2                        & 0.630       & 0.785        \\ \midrule
    4                        & \textbf{0.645 }      & \textbf{0.799}        \\
 \bottomrule
  \end{tabular}
  \label{tab:modality_attention}
  \vspace{-1.2em}
\end{table}




\vspace{-0.3em}
\subsection{Visualization}




To further validate the interpretability and fine-grained temporal localisation capabilities of our MultiHateLoc framework, we present two representative case studies from the HateMM dataset in Fig.~\ref{fig:case_study_visualization}. Each case is visualised using (i) a timeline showing the frame-level hate prediction probability, (ii) sampled video frames, (iii) aligned audio waveform segments, and (iv) corresponding transcribed text. These qualitative examples demonstrate that our model not only accurately performs weakly-supervised hate localisation but also adapts to the most informative modality at each timestamp.

In the first case (top row), the majority of the video content is non-hateful; however, several hate frames are briefly inserted. Our model successfully identifies these short hate segments, as shown by the sharply rising prediction scores and the correctly highlighted frames. This example demonstrates the model's sensitivity to short-term yet salient visual cues and its ability to detect hate signals even when they are sparsely distributed over time.

In contrast, the second case (bottom row) contains a visually benign video stream (a casual video call), while hateful expressions appear in both the audio and text modalities. Despite the absence of visual hate cues, MultiHateLoc accurately identifies the hate segment via textual and auditory clues. The model activates appropriately in response to hateful words or phrases, as evident from the red-highlighted audio waveform peaks and offensive transcriptions. This showcases our model’s capability to rely on non-visual modalities when appropriate, underlining the strength of our dynamic modality selection and cross-modal attention mechanisms.

Together, these examples highlight the robustness and interpretability of MultiHateLoc: it not only localises hate at the frame level, but also attributes it to the correct modality, even in challenging cases with implicit or modality-specific expressions of hate.

\section{Conclusion}

We introduced MultiHateLoc, the first framework for weakly supervised multimodal hate content localisation, addressing a task that has been largely overlooked in prior research. By combining dynamic modality selection, cross-modal attention, and a modality-aware MIL objective, MultiHateLoc effectively captures the sparse and asynchronous cues that characterise hateful content across video, audio, and text.
Experiments demonstrate that MultiHateLoc can produce accurate temporal localisation under video-level supervision. 
Comprehensive ablations verify the contribution of each component, and qualitative visualisations further show how the model identifies discriminative segments across modalities.
Overall, this work opens a new research direction for fine-grained multimodal hateful video understanding under weak supervision. 



\vspace{-0.5em}
\section{Acknowledgments}
This work was supported by the Alan Turing Institute and DSO National Laboratories Framework Grant Funding.

\balance
\bibliographystyle{ACM-Reference-Format}
\bibliography{sample-base}

\appendix

\end{document}